\documentclass[runningheads,a4paper]{llncs}

\usepackage[american]{babel}
\usepackage[%
rm={oldstyle=false,proportional=true},%
sf={oldstyle=false,proportional=true},%
tt={oldstyle=false,proportional=true,variable=true},%
qt=false%
]{cfr-lm}
\usepackage{graphicx}
\usepackage{csquotes}
\usepackage[backend=bibtex,style=ieee,sorting=ynt]{biblatex}
\begin{document} 

\title{Convolutional Radio Modulation Recognition Networks}

\author{Timothy J. O'Shea \inst{1} \and Johnathan Corgan\inst{2} \and T. Charles Clancy \inst{1} }

\institute{
 Bradley Department of Electrical and Computer Engineering, Virginia Tech, 900 N Glebe Road, Arlington, VA 22203 USA
\email{oshea@vt.edu}
\and
Corgan Labs, 6081 Meridian Ave., Suite 70-111, San Jose, CA 95120
\email{johnathan@corganlabs.com}
}

\maketitle

\begin{abstract} 
We study the adaptation of convolutional neural networks to the complex-valued temporal radio signal domain.  We compare the efficacy of radio modulation classification using naively learned features against using expert feature based methods which are widely used today and e show significant performance improvements.  We show that blind temporal learning on large and densely encoded time series using deep convolutional neural networks is viable and a strong candidate approach for this task especially at low signal to noise ratio.
\end{abstract} 


\section{Introduction}
Radio communications present a unique signal processing domain with a number of interesting challenges and opportunities for the machine learning community.
In this field expert features and decision criterion have been extensively developed, and analyzed for optimality under specific criteria for many years.  
However in the past few years the trend in machine learning applied to image processing \cite{imagenet} and voice recognition \cite{cldnn} is overwhelmingly that of feature learning from data rather than crafting of expert features, suggesting we should evaluate a similar shift in this domain.

Concurrently wireless data demand is driving a need for improved radio efficiency.  High quality spectrum sensing and adaptation to improve spectral allocation and interference mitigation is an important route by which we may achieve this.  The FCC in the United States as well as counterparts in Europe are taking seriously and pursuing spectrum policy which leverages some of this ideas from Dynamic Spectrum Access (DSA)\cite{dsa}, making clear the need for improved spectrum sensing and signal identification algorithms allowing sensors and radios to detect and identify spectrum users and interferers at the best possible range and thus signal to noise ratio.

Ideas such as DSA, opportunistic access and sharing of spectrum, and "Cognitive Radio" (CR) \cite{cograd}, a more broad class of radio optimization through learning, have been widely discussed at the conceptual level.  Efforts in these fields however have been constrained to relatively specialized solutions which lack the generality needed to deal with a complex and growing number emitter types, interference types and propagation environments.   \cite{cumcr} \cite{mlcrn} \cite{rondeau}

This is a significant challenge in the community as expert systems designed to perform well on specialized tasks often lack flexibility and can be expensive and tedious to develop analytically.

Building upon successful strategies from image and voice recognition domains in machine learning, we demonstrate an approach in the radio using Convolutional Neural Networks (CNNs) and Deep Neural Networks (DNNs) which offers flexibility to learn features across a wide range of tasks and demonstrates improved classification accuracy against current day approaches.

\section{Modulation Recognition}
In Dynamic Spectrum Access (DSA) one of the key sensing performed is that of providing awareness of nearby emitters to avoid radio interference and optimize spectrum allocation.  This identifying and differentiating broadcast radio, local and wide area data and voice radios, radar users, and other sources of potential radio interference in the vicinity which each have different behaviors and requirements.  Modulation Recognition then is the task of classifying the modulation type of a received radio signal as a step towards understanding what type of communications scheme and emitter is present.   

This can be treated as an N-class decision problem where our input is a complex base-band time series representation of the received signal.  That is, we sample in-phase and quadrature components of a radio signal at discrete time steps through an analog to digital converted with a carrier frequency roughly centered on the carrier of interest to obtain a 1xN complex valued vector.  Classically, this is written as in equation \ref{eq:classic} where $s(t)$ is a time series signal of either a continuous signal or a series of discrete bits modulated onto a sinusoid with either varying frequency, phase, amplitude, trajectory, or some permutation of multiple thereof.  $c$ is some path loss or constant gain term on the signal, and $n(t)$ is an additive Gaussian white noise process reflecting thermal noise.

\begin{equation} \label{eq:classic}
r(t) = s(t)*c + n(t)
\end{equation}

Analytically, this simplified expression is used widely in the development of expert features and decision statistics, but the real world relationship looks much more like that given in equation \ref{eq:realistic} in many systems.

\begin{equation} \label{eq:realistic}
r(t) = e^{j*n_{Lo}(t)} \int_{\tau=0}^{\tau_0} s(n_{Clk}(t-\tau))h(\tau) + n_{Add}(t) 
\end{equation}

This considers a number of real world effects which are non-trivial to model: modulation by a residual carrier random walk process, $n_{Lo}(t)$, resampling by a residual clock oscillator random walk, $n_{Clk}(t)$, convolution with a time varying rotating non-constant amplitude channel impulse response $h(t-\tau)$, and the addition of noise which may not be white, $n_{Add}(t)$.   Each presents an unknown time varying source of error.

Modeling expert features and decision metrics optimality analytically under each of these harsh realistic assumptions on propagation effects is non-trivial and often forces simplifying assumptions.  In this paper we focus on empirical measurement of performance in harsh simulated propagation environments which include all of the above mentioned effects, but do not attempt to analytically trace their performance in closed form.

\subsection{Expert Cyclic-Moment Features}

Integrated cyclic-moment based features \cite{cumulant} are currently widely popular in performing modulation recognition and for forming analytically derived decision trees to sort modulations into different classes.   In general, they take the form given in equation \ref{eq:cyc}.

\begin{equation} \label{eq:cyc}
s_{nm} = f_m( x^n(t) ... x^n(t+T) )
\end{equation}

By computing the m'th order statistic on the n'th power of the instantaneous or time delayed received signal r(t), we may obtain a set of statistics which uniquely separate it from other modulations given a decision process on the features.  For our expert feature set, we compute 32 features.  These consist of cyclic time lags of 0 and 8 samples.  And the first 2 moments of the first 2 powers of the complex received signal, the amplitude, the phase, and the absolute value of phase for each of these lags.

We train several classifiers on these set of expert features as a benchmark comparison.  These leverage scikit-learn and consist of a Decision Tree, K=1-Nearest Neighbor, Gaussian Naive Bayes, and an RBF-SVM.  Additionally, we train a 3-layer deep neural network consisting only of fully connected layers of size 512, 256, and 11 neurons.  Each of these is measured to provide a performance baseline  estimate for how a current day system operating on such a feature set might perform.  Best expert-feature performance is obtained from the SVM and DNN based approaches.

\subsection{Convolutional Feature Learning}

We evaluate several feature learning methods, but our principal method is that of a convolutional neural network (CNN) provided with a windowed input of the raw radio time series $r(t)$.  We treat the complex valued input as an input dimension of 2 real valued inputs and use $r(t)$ as a set of 2xN vectors into a narrow 2D Convolutional Network where the orthogonal synchronously sampled In-Phase and Quadrature (I \& Q) samples make up this 2-wide dimension.

\section{Evaluation Dataset}

While simulation and the use of synthetic data sets for learning is sometimes frowned upon in machine learning, radio communications presents a special case.  Training with real data is important and valuable - and will be addressed in future work - but certain properties of the domain allow us to say our simulation is quite meaningful.

Radio communications signals are in reality synthetically generated, and we do so deterministically in a way identical to a real system, introducing modulation, pulse shaping, carried data, and other well characterized transmit parameters identical to a real world signal.   We modulate real voice and text data sets onto the signal.   In the case of digital modulation we whiten the data using a block randomizer to ensure bits are equiprobable.

Radio channel effects are relatively well characterized. We employ robust models for time varying multi-path fading of the channel impulse response, random walk drifting of carrier frequency oscillator and sample time clocks, and additive Gaussian white noise.  We pass our synthetic signal sets through harsh channel models which introduce unknown scale, translation, dilation, and impulsive noise onto our signal.

We model the generation of this dataset in GNU Radio \cite{gnuradio} using the GNU Radio channel model \cite{channelmodel} blocks and then slice each time series signal up into a test and traning set using a 128 samples rectangular windowing process.  The total dataset is roughly 500 MBytes stored as a python pickle file with complex 32 bit floating point samples.

\subsection{Dataset Availability}

This data will hopefully be of great use to others in the field and may serve as a benchmark for this domain.  This dataset is available in pickled python format at \url{http://radioml.com}, consisting of time-windowed examples and corresponding modulation class and SNR labels.  We hope to grow scope of modulations addressed and the channel realism as interest in this area.

\subsection{Dataset Parameters}

We focus on a dataset consisting of 11 modulations: 8 digital and 3 analog modulation, all are widely used in wireless communications systems all around us.   These consist of BPSK, QPSK, 8PSK, 16QAM, 64QAM, BFSK, CPFSK, and PAM4 for digital modulations, and WB-FM, AM-SSB, and AM-DSB for analog modulations.  Data is modulated at a rate of roughly 8 samples per symbol with a normalized average transmit power of 0dB.

\subsection{Dataset Visualization}

\begin{figure}[ht]
\centering
\begin{minipage}[b]{0.45\linewidth}
      \includegraphics[width=1.0\textwidth]{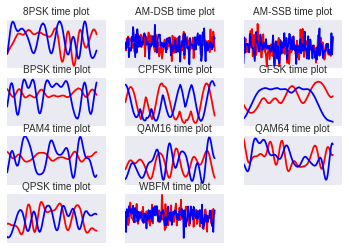}
  \caption{Time Domain of High-SNR Example Classes}
\label{fig:timedomain}
\end{minipage}
\quad
\begin{minipage}[b]{0.45\linewidth}
      \includegraphics[width=1.0\textwidth]{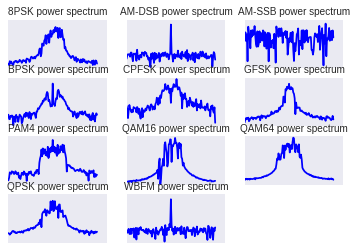}
  \caption{Power Spectrum of High-SNR Example Classes}
\label{fig:powerspectrum}
\end{minipage}
\end{figure}

Inspecting a single example from each class of modulation in the time (figure \ref{fig:timedomain}) and frequency domain (figure \ref{fig:powerspectrum}), we see a number of similarities and differences between modulations visually, but due to pulse shaping, distortion and other channel effects they are not all readily discernible by a human expert visually. 

In the frequency domain, each of the signals follows a similar band-limited power envelope by design whose shape provides some clues as to the modulation, but against poses a difficult noisy task for a human expert to judge visually.

\subsection{Modulated Information}

In radio communications, signals are typically comprised of a number of modulated data bits on well defined and understood basis functions into discrete modes formed by these bases.  Complex baseband representation of a signal decomposes a radio voltage level time-series into its projections onto the sine and cosine functions at a carrier frequency.  By manipulating the frequency, amplitude, phase, or sum thereof data bits are then modulated into this space through discrete and separable modes for each distinct symbol period in time in the case of digital, or continuous location in the case of analog modulation.  For the case of QPSK this phase-mapping is shown in \ref{eq:qpsk}.

\begin{equation}\label{eq:qpsk}
s(t_i) = e^{j 2 \pi f_c t + \pi \frac{2 c_i + 1}{4}}, c_i \in {0, 1, 2, 3}
\end{equation}

Pulse shaping filters such as root-raised cosine are then typically applied to band-limit the signal in frequency and remove sharp wide-band transients between these distinct modes, resulting in mixing of adjacent symbols' bases at the transmitter in a deterministic and invertible.   In our simulated data set we use a root-raised cosine pulse shaping filter with an excess bandwidth of 0.35 for each digital signal.

\subsection{Effects on the Modulated Signal}

Channel effects in contrast are not deterministic and not completely invertible in a communications system.   Real systems experience a number of effects on the transmitted signal, which make recovery and representation thereof challenging.  Thermal noise results in relatively flat white Gaussian noise at the receiver which forms a noise floor or sensitivity level and signal to noise ratio.  Oscillator drift due to temperature and other semiconductor physics differing at the transmitter and receiver result in symbol timing offset, sample rate offset, carrier frequency offset and phase difference.  These effects result in a temporal shifting, scaling, linear mixing/rotating between channels, and spinning of the received signal based on unknown time varying processes.   Last, real channels undergo random filtering based on the arriving modes of the transmitted signal at the receiver with varying amplitude, phase, Doppler, and delay.  This is a phenomenon commonly known as multi-path fading or frequency selective fading, which occurs in any environment where signals may reflect off buildings, vehicles, or any form of reflector in the environment.  This is typically removed at the receiver by the estimation of the instantaneous value of the time varying channel response and deconvolution of it from the received signal.

\subsection{Generating a dataset}

To generate a well characterized dataset, we select a collection of modulations which are used widely in practice and operate on both discrete binary alphabets (digital modulations), and continuous alphabets (analog modulations).  We modulate known data over each modem and expose them each to the channel effects described above using GNU Radio.   We segment the millions of samples into a dataset consisting of numerous short-time windows in a fashion similar to how a continuous acoustic voice signal is typically windowed for voice recognition tasks.   We extract steps of 128 samples with a shift of 64 samples to form our extracted dataset.

After segmentation, examples are roughly 128 $\mu$ sec each assuming a sample rate of roughly 1 MSamp/sec.   Each contains between 8 and 16 symbols with random time offset, scaling, rotation, phase, channel response, and noise.   These examples represent information about the modulated data bits, information about how they were modulated, information about the channel effects the signal passed through during propagation, and information about the state of the transmitted and receiver device states and contained random processes.   We focus specifically on recovering the information about how the signal was modulated and thus label the dataset according to a discrete set of 11 class labels corresponding to the modulation scheme.

\section{Technical Approach}

In a radio communication system, one class of receiver which is commonly considered is a "matched-filter" receiver.  That is on the receive side of a communications link, expert designed filters matched with each transmitted symbol representation are convolved with the incoming time signal, and form peaks as the correct symbol slides over the correct symbol time in the received signal.   By convolving, we average out the impulsive noise in the receiver in an attempt to optimize signal to noise.  Typically, before this convolutional stage, symbol timing and carrier frequency is recovered using an expert envelope or moment based estimators derived analytically for a specific modulation and channel model.   The intuition behind the use of a convolutional neural network in this application then is that they will learn to form matched filters for numerous temporal features, each of which will have some filter gain to operate at lower SNR, and which when taken together can form a robust basis for classification.

\subsection{Learning Invariance}

Many of these recovery processes in radio communications systems can be thought of in terms of invariance to linear mixing, rotation, time shifting, scaling, and convolution through random filters (with well characterized probabilistic envelopes and coherence times).   These are analogous to similar learning invariance which is heavily addressed in vision domain learning where matched filters for specific items or features in the image may undergo scaling, shifting, rotation, occlusion, lighting variation, and other forms of noise.   We seek to leverage the shift-invariant properties of the convolutional neural network to be able to learn matched filters which may delineate symbol encoding features naively, without expert understanding or estimation of the underlying waveform.  

\subsection{Evaluation Networks}

\begin{figure}[ht]
\centering
\begin{minipage}[b]{0.45\linewidth}
      \includegraphics[width=1\textwidth]{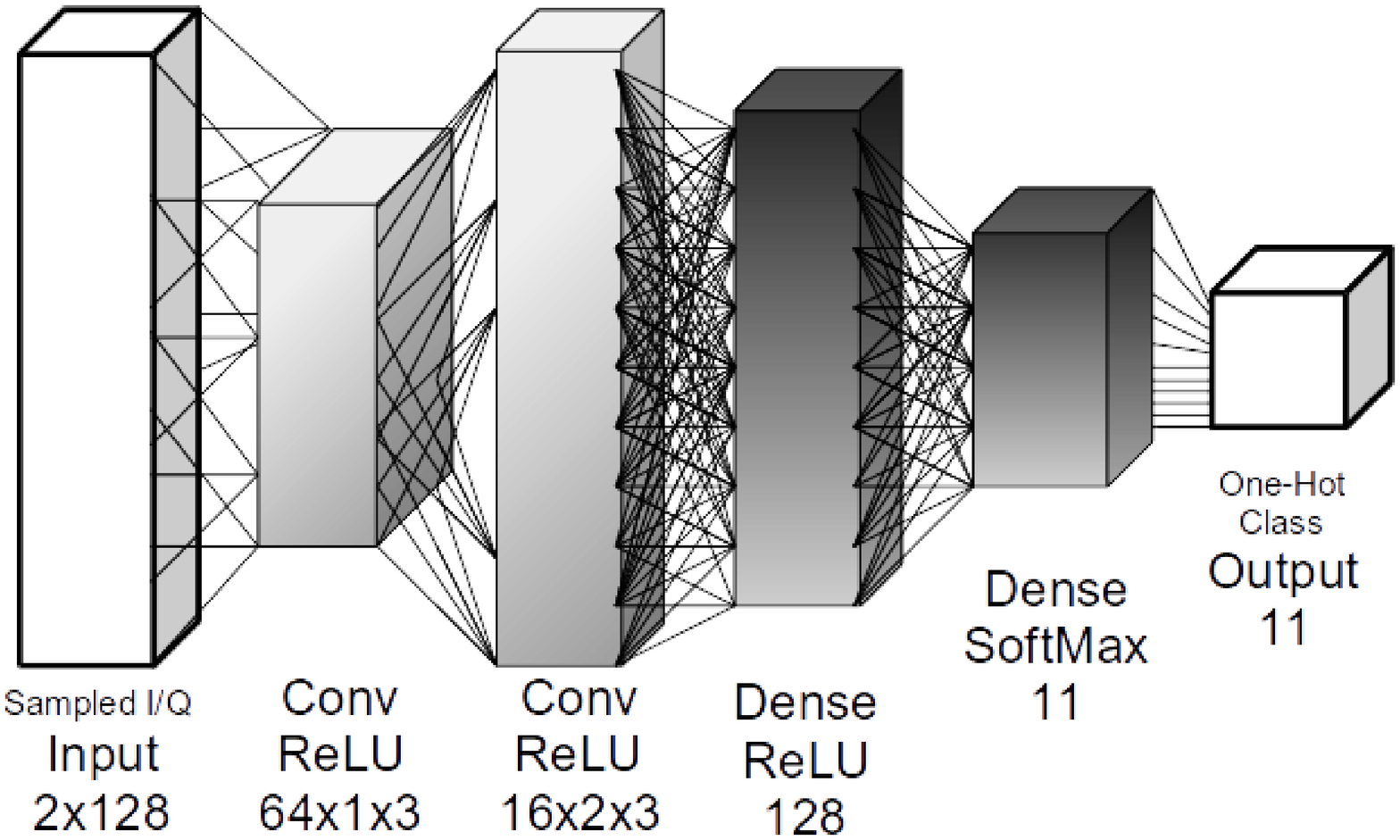}
  \caption{CNN Architecture}\label{figure:network}
\end{minipage}
\quad
\begin{minipage}[b]{0.45\linewidth}
      \includegraphics[width=1\textwidth]{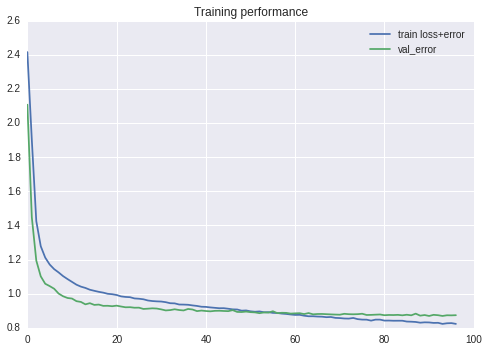}
  \caption{Loss Plots CNN2 60\% Dropout}\label{figure:loss}
\end{minipage}
\end{figure}

We train against several candidate neural networks.  A 4-layer network utilizing two convolutional layers and two dense fully connected layers (CNN and CNN2).  Layers use rectified linear (ReLU) activation functions except for a Softmax activation on the one-hot output layer.  We use this network depth as it is roughly equivalent to networks which work well on similar simple datasets in the vision domain such as MNIST.

Regularization is used to prevent over-fitting. CNN uses Dropout, a $ \left \| W \right \|_2 $ norm penalty on the convolutional layer weights, encouraging minimum energy bases, and a $ \left \| \mathbf{h} \right \|_1 $ norm penalty on the first dense layer activation, to encourage sparsity of solutions \cite{l1l2reg} \cite{deconv}.  CNN2 uses only dropout, and DNN uses only dropout.
Training is conducted using a categorical cross entropy loss function and an Adam \cite{adam} solver which seems to slightly outperform RMSProp \cite{rmsprop} on our dataset.  We implement our network training and prediction in Keras \cite{keras} running on top of TensorFlow \cite{tensorflow} on an NVIDIA Cuda \cite{cuda} enabled Titan X GPU in a DIGITS Devbox.

An illustration of the CNN architecture is shown in figure \ref{figure:network}.  CNN2 is identical but larger, containing 256 and 80 filters in layers 1 and 2, and 256 neurons in layer 3.  The DNN evaluated contains 4 dense layers of size 512, 256, 128, and n-classes neurons.

\subsection{Training Complexity}

We train our highest complexity model for approximately 23 minutes with the Adam solver over the $\sim 900,000$ sample training set in batch sizes of 1024.  Epochs take roughly 15 seconds and we do observe some over-fitting despite out regularization, but validation loss does not significantly inflect and we keep the best validation loss model for evaluation.  
\ref{figure:loss}.   

\subsection{Learned Features}

\begin{figure}[ht]
\centering
\begin{minipage}[b]{0.45\linewidth}
      \includegraphics[width=1\textwidth]{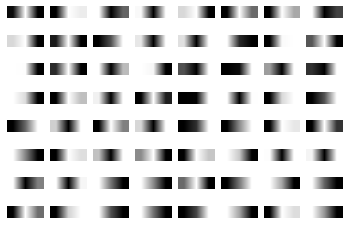}
  \caption{Conv1 Layer Weights (1x3 Filters)}
  \label{figure:w1}
\end{minipage}
\quad
\begin{minipage}[b]{0.45\linewidth}
      \includegraphics[width=1\textwidth]{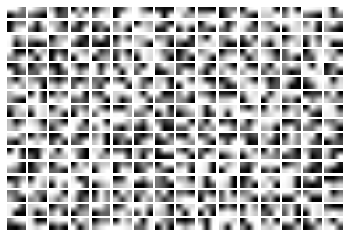}
  \caption{Conv2 Layer Weights (2x3 Filters)}\label{figure:w2}
\label{fig:minipage2}
\end{minipage}
\end{figure}

Plotting learned features can sometimes give us an intuition as to what the network is learning about the underlying representation.  In this case, we plot convolutional layer 1 and convolutional layer 2 filter weights below.   In figure \ref{figure:w1}, the first layer, we have 64 filters of 1x3.  In this case we simply get a set of 1D edge and gradient detectors which operate across each I and the Q channel.

In convolutional layer 2, weights shown in figure \ref{figure:w2} we compose this first layer feature map into 64*16x2x3 larger feature maps, which comprise what is occurring on both I and Q channels simultaneously.  These feature maps do not look hugely different than those seen at the lower levels of an image conv-net comprising of 2D learned edge detectors and Gabor-like filters.   

\section{Results}

To evaluate the performance of our classifier, we look at classification performance on a test data set.  We train on a corpus of approximately 12 million complex samples divided across the 11 modulations.  These are divided into training examples of 128 samples in length.   We use approximately 96,000 example for training, and 64,000 examples for testing and validation.   These samples are uniformly distributed in SNR from -20dB to +20dB and tagged so that we can evaluate performance on specific subsets.

After training, we achieve roughly a 87.4\% classification accuracy across all signal to noise ratios on the test dataset, but to understand the meaning of this we must inspect how this classification accuracy breaks down across the SNR values of the different training examples, and how it compares to the performance of existing expert feature based based classifiers.

\begin{figure}[ht!]
  \centering
      \includegraphics[width=1.0\textwidth]{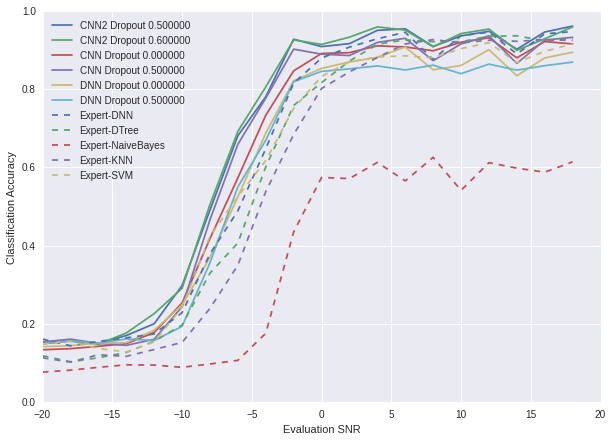}
  \caption{Classifier Performance vs SNR} \label{figure:modreq_snr}
\end{figure}

Plot test set modulation classification accuracy as a function of example signal to noise ratio for each classifier in\ref{figure:modreq_snr}.  Solid lines show classifiers trained directly on the radio time series data performing deep feature learning, while dotted lines indicate classifiers using only the expert features previously described as input.
This view is a critical way to inspect results as performance at low SNR impacts range and coverage area over which we can effectively use the classifier.
We obtain significantly better low-SNR classification accuracy performance from large convolutional neural networks (CNN2) with significant amounts of dropout regularization (0.6).  At low-SNR the best CNN model is outperforming expert feature based systems by 2.5-5dB of SNR, while after +5dB SNR performance is similar.  This is a significant performance improvement, and one that could potentially at least double effective coverage area of a sensing system.

\begin{figure}[ht!]
  \centering
      \includegraphics[width=0.7\textwidth]{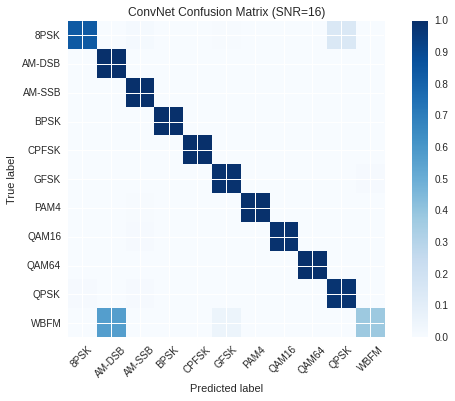}
  \caption{Conv Net Confusion Matrix at +18dB SNR}\label{figure:bestconf}
\end{figure}

For our highest SNR case CNN2(0.6) classification we show a confusion matrix in figure \ref{figure:bestconf}.   At +18dB SNR, we have a clean diagonal in the confusion matrix and can see our remaining discrepancies are that of 8PSK misclassified as QPSK, and WBFM misclassified as AM-DSB.   Both of these are explainable in the underlying dataset.  An 8PSK symbol containing the specific bits is indiscernible from QPSK since the QPSK constellation points are spanned by 8PSK points.  In the case of WBFM/AM-DSB the analog voice signal has periods of silence where only a carrier tone is present making these examples indiscernible.  Therefore it is unlikely that and accuracy of 100\% can be obtained even at high SNR on this data set and making the remaning confusion reasonably tolerated.

To get a better understanding of how performance varies with SNR, we inspect confusion matrices for several classifiers at differing SNR levels.

\begin{figure}[ht]
\centering
\begin{minipage}[b]{0.45\linewidth}
      \includegraphics[width=1\textwidth]{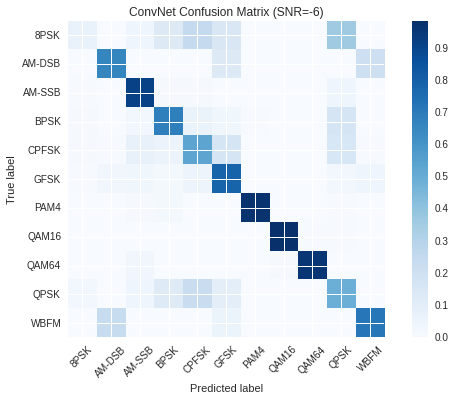}
  \caption{-6dB Performance of CNN2 on Raw Sample Data}\label{figure:vlow1}
\end{minipage}
\quad
\begin{minipage}[b]{0.45\linewidth}
      \includegraphics[width=1\textwidth]{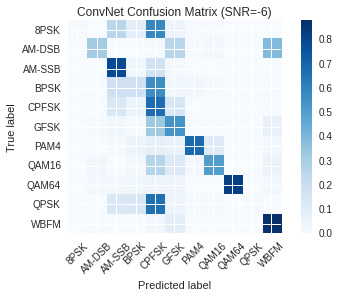}
  \caption{-6dB Performance of DNN on Expert Features}\label{figure:vlow2}
\end{minipage}
\end{figure}

\begin{figure}[ht]
\centering
\begin{minipage}[b]{0.45\linewidth}
      \includegraphics[width=1\textwidth]{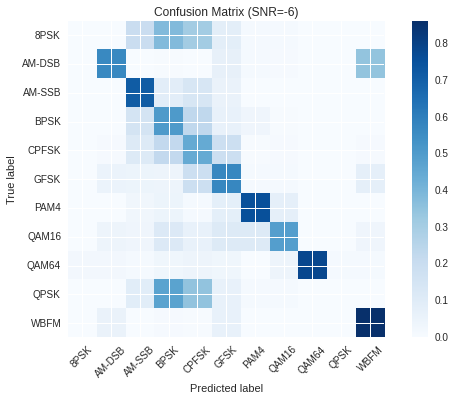}
  \caption{-6dB Performance of SVM on Expert Features}\label{figure:vlow3}
\end{minipage}
\quad
\begin{minipage}[b]{0.45\linewidth}
      \includegraphics[width=1\textwidth]{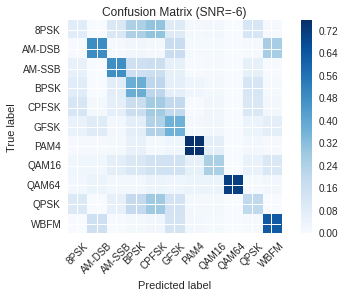}
  \caption{-6dB Performance of Decision Tree on Expert Features}\label{figure:vlow4}
\end{minipage}
\end{figure}

At very low SNR (-6dB), in figure \ref{figure:vlow1}, \ref{figure:vlow2}, \ref{figure:vlow3}, and \ref{figure:vlow4} we see an interesting case where all are around 50\% accuracy within +-20\%.  In this case the cleaner diagonal on the CNN2 classifier is significantly more pronounced than the other 3 cases shown, in this region learned features have a significant performance advantage.

\begin{figure}[ht]
\centering
\begin{minipage}[b]{0.45\linewidth}
      \includegraphics[width=1.0\textwidth]{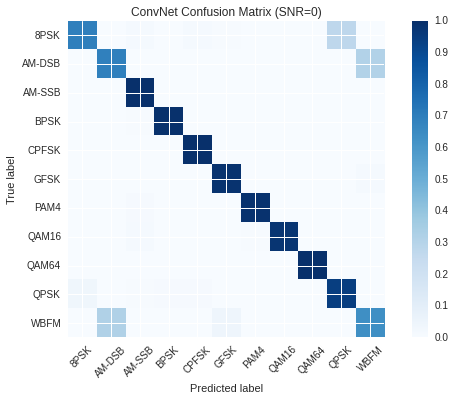}
  \caption{0dB Performance of CNN2 on Raw Sample Data}\label{figure:low1}
\end{minipage}
\quad
\begin{minipage}[b]{0.45\linewidth}
      \includegraphics[width=1.0\textwidth]{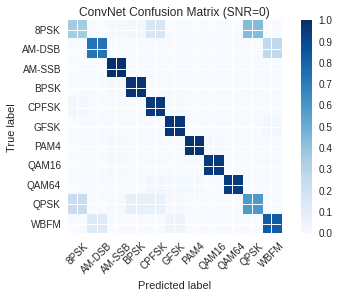}
  \caption{0dB Performance of DNN on Expert Features}\label{figure:low2}
\end{minipage}
\end{figure}

\begin{figure}[ht]
\centering
\begin{minipage}[b]{0.45\linewidth}
      \includegraphics[width=1.0\textwidth]{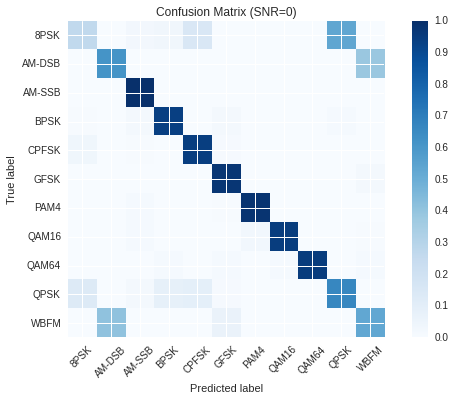}
  \caption{0dB Performance of SVM on Expert Features}\label{figure:low3}
\end{minipage}
\quad
\begin{minipage}[b]{0.45\linewidth}
      \includegraphics[width=1.0\textwidth]{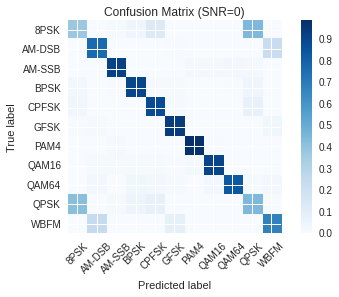}
  \caption{0dB Performance of Decision Tree on Expert Features}\label{figure:low4}
\end{minipage}
\end{figure}

At slightly higher, but still low SNR (0dB) performance for all 4 classifiers now has a well defined diagonal, but we see less mis-classifications occurring off-diagonal, especially in the 8PSK case.

\section{Model Complexity}

An important consideration in many radio system is the training and classification run time due to computational complexity.  One common critique of deep learning is its need for large amounts of compute resources, however in this paper our network is relatively compact and the dataset relatively small.  We compare the training and classification run times for each of the models below.

\begin{figure}[ht]
\centering
\begin{minipage}[b]{0.45\linewidth}
      \includegraphics[width=1\textwidth]{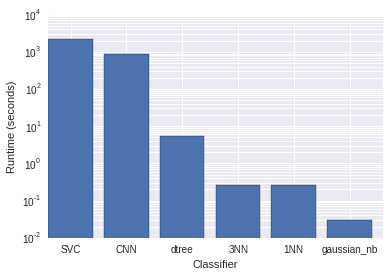}
  \caption{Model Training Runtime in Seconds}\label{figure:traintime}
\end{minipage}
\quad
\begin{minipage}[b]{0.45\linewidth}
      \includegraphics[width=1\textwidth]{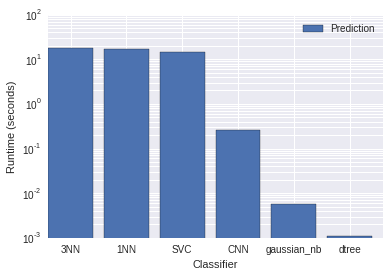}
  \caption{Signal Classification Time in Seconds (per SNR-batch)}\label{figure:exectime}
\end{minipage}
\end{figure}

In figure \ref{figure:traintime} we can see that our CNN model does take a significant amount of time to train, but is lower than the time required for for the SVM training case.

In figure \ref{figure:exectime} it turns out that classification time with this model using Keras compiled python is significantly faster than most of the other models including nearest-neighbor and SVM models using scikit-learn.  Only the Decision Tree and GaussianNB models obtain faster classification run times.

In both cases, a ConvNet based classification model of this scale for such a dataset presents an attractive choice for this task when classification performance is considered.

\section{Conclusions}

While these results are not a comprehensive comparison of existing best case expert feature based modulation classifiers, they do demonstrate that compared to a relatively well expert regarded approach, blind Convolutional Networks on time series radio signal data are viable and work quite well.  In figure \ref{figure:modreq_snr}, we compared accuracy to SNR for several classifier strategies and believe that for low SNR and short-time examples (128 complex samples), this represents a powerful and likely state of the art accuracy approach to modulation classification.  This approach holds the potential to easily scale to additional modulation classes and should be considered as a strong candidate for DSA and CR systems which rely on robust low SNR classification of radio emitters.

\section{Future Work}

Our results compare to a reasonable approximation of the current best expert system approach, but because no robust competition data sets exist in the emerging field of machine learning in the radio domain, it is difficult to directly compare performance to current state of the art approaches.   We hope to further evaluate this in later work, and improve both the feature learning and expert approaches from their current level.
Performance refinements are inevitable on the CNN2 network architecture, we expended some effort optimizing it but did not do so exhaustively.  Larger filters, differing architecture, and pooling layers all may affect performance significantly, but were not fully investigated for their suitability in this work.
Numerous additional techniques could be applied to the problem including the introduction of invariance to additional channel induced effects such as dilation, I/Q imbalance, phase offset and others.  Spatial Transformer Networks \cite{stn} have demonstrated a powerful ability to learn this type of invariance on image data and may serve as an interesting candidate for enabling improved invariance learning to these effects.
Sequence models and recurrent layers \cite{speechseq} may be able to represent a signal sequence embedding and will almost certainly prove valuable in longer time representation, but we have yet to investigate this area fully.
This application domain is ripe for a wide array of further investigation and applications which will significantly impact the state of the art in wireless signal processing and cognitive radio domains, shifting them more towards machine learning and data driven approaches.

\section*{Acknowledgments}

The authors would like to thank the Bradley Department of Electrical and Computer Engineering at the Virginia Polytechnic Institute and State University, the Hume Center, and DARPA all for their generous support in this work.

This research was developed with funding from the Defense Advanced Research Projects Agency's (DARPA) MTO Office under grant HR0011-16-1-0002. The views, opinions, and/or findings expressed are those of the author and should not be interpreted as representing the official views or policies of the Department of Defense or the U.S. Government.
 
 
\printbibliography

\end{document}